# Structural Similarity Index SSIMplified:

*Is there really a simpler concept at the heart of image quality measurement?*

Kieran G. Larkin

**Nontrivialzeros Research**

*May 2015*

The Structural Similarity Index (SSIM) is generally considered to be a milestone in the recent history of Image Quality Assessment (IQA). Alas, SSIM's accepted development from the product of three heuristic factors continues to obscure it's real underlying simplicity.

Starting instead from a symmetric-antisymmetric reformulation we first show SSIM to be a contrast or visibility function in the classic sense. Furthermore, the previously enigmatic structural covariance is revealed to be the difference of variances. The second step, eliminating the intrinsic quadratic nature of SSIM, allows a near linear correlation with human observer scores, and without invoking the usual, but arbitrary, sigmoid model fitting. We conclude that SSIM can be re-interpreted in terms of perceptual masking: it is essentially equivalent to a normalised error or noise visibility function (NVF), and, furthermore, the NVF alone explains it success in modelling perceptual image quality. We use the term Dissimilarity Quotient (DQ) for the specifically anti/symmetric SSIM derived NVF.

It seems that IQA researchers may now have two choices:

- *Continue to use the complex SSIM formula, but noting that SSIM only works coincidentally since the covariance term is actually the mean square error (MSE) in disguise.*
- *Use the simplest of all perceptually-masked image quality metrics, namely NVF or DQ.*

On this choice Occam is clear: in the absence of differences in predictive ability, the fewer assumptions that are made, the better.

## Symmetric Reformulation

A few simple substitutions dramatically simplify the SSIM of Wang and Bovik [1]. Because of the partial cancellation of the variance and covariance terms, the SSIM for two images S can be written as the product of 2 partial indices, $S_L$ and $S_V$:

$$\text{SSIM}\{f_1, f_2\} = S = \left[\frac{2\mu_1\mu_2}{\mu_1^2 + \mu_2^2}\right] \times \left[\frac{2\sigma_1\sigma_2}{\sigma_1^2 + \sigma_2^2}\right] \times \left[\frac{\sigma_{12}}{\sigma_1\sigma_2}\right]$$

$$S = \left[\frac{2\mu_1\mu_2}{\mu_1^2 + \mu_2^2}\right] \times \left[\frac{2\sigma_{12}}{\sigma_1^2 + \sigma_2^2}\right] = S_L \cdot S_V \quad (1)$$

The first index $S_L$ contains local luminance and the second $S_V$ local covariance/variance.

## SSIM symmetry/antisymmetry

Observe that SSIM is a symmetric measure relative to image ordering $f_1 \leftrightarrow f_2$, so that it seems natural to formulate in terms of symmetric/antisymmetric image constructs:

$$\begin{aligned} f_+(\mathbf{x}) &= f_2(\mathbf{x}) + f_1(\mathbf{x}) & 2f_1(\mathbf{x}) &= f_+(\mathbf{x}) - f_-(\mathbf{x}) \\ f_-(\mathbf{x}) &= f_2(\mathbf{x}) - f_1(\mathbf{x}) & 2f_2(\mathbf{x}) &= f_+(\mathbf{x}) + f_-(\mathbf{x}) \end{aligned} \quad (2)$$

These can be interpreted simply as the (even) sum image $f_+$ and the (odd) difference $f_-$ image respectively. Now introduce the means μ and variances σ for the even and odd images in exactly the same way as for original images. The local statistics are straightforward and covered in all the conventional texts, resulting in:

$$\mu_+ = \mu_2 + \mu_1, \quad \mu_- = \mu_2 - \mu_1$$

$$\left.\begin{aligned} 4\mu_1\mu_2 &= \mu_+^2 - \mu_-^2 \\ 2\mu_1^2 + 2\mu_2^2 &= \mu_+^2 + \mu_-^2 \end{aligned}\right\} \quad \left.\begin{aligned} 4\sigma_{12} &= \sigma_+^2 - \sigma_-^2 \\ 2\sigma_1^2 + 2\sigma_2^2 &= \sigma_+^2 + \sigma_-^2 \end{aligned}\right\} \quad (3)$$

Which means the (anti)symmetrical image statistics are simply related to the original image statistics. Crucially the numerators can be represented as the difference of squares; numerators as the sum of squares. Accordingly the SSIM now has a covariance free formulation: the terms on the RHS can be interpreted directly as a (squared) luminance contrast and a variance contrast (in the sense of a Michelson contrast or visibility as described by Peli [2].

$$\text{SSIM}\{f_1, f_2\} = \frac{2\mu_1\mu_2}{\mu_1^2 + \mu_2^2} \cdot \frac{2\sigma_{12}}{\sigma_1^2 + \sigma_2^2} = \frac{(\mu_+^2 - \mu_-^2)}{(\mu_+^2 + \mu_-^2)} \cdot \frac{(\sigma_+^2 - \sigma_-^2)}{(\sigma_+^2 + \sigma_-^2)} \quad (4)$$

Many important types of image distortion (jpeg compression, additive noise, blurring etc) induce negligible variation in the luminance term, and we ignore it henceforward:

$$\left.\begin{array}{l}\mu_o \simeq 0 \\ \mu_1 \simeq \mu_2 \simeq \mu_e\end{array}\right\} \Rightarrow \frac{2\mu_1\mu_2}{\mu_1^2 + \mu_2^2} \simeq 1, \quad \text{SSIM}\{f_1, f_2\} \simeq S_V = \frac{2\sigma_{12}}{\sigma_1^2 + \sigma_2^2} = \frac{(\sigma_+^2 - \sigma_-^2)}{(\sigma_+^2 + \sigma_-^2)} \tag{5}$$

Applying the linear operations of negation and offset to SSIM [3] gives the classic form of a normalised dissimilarity index (RHS):

$$1 - \text{SSIM}\{f_1, f_2\} = 1 - \frac{(\sigma_+^2 - \sigma_-^2)}{(\sigma_+^2 + \sigma_-^2)} = \frac{2\sigma_-^2}{\sigma_+^2 + \sigma_-^2} \tag{7}$$

In most published comparisons of the SSIM versus human opinion (MOS and DMOS) the highly curved scatterplots are straightened up with a heuristic logistic curve remapping. Yet it seems a trivial square root operation can achieve much the same effect:

$$\sqrt{1 - \text{SSIM}\{f_1, f_2\}} = \frac{\sqrt{2}\sigma_-}{\sqrt{\sigma_+^2 + \sigma_-^2}} \tag{8}$$

The above formulation is ubiquitous: for example the Divisive Normalisation of Laparra's [4] image quality measure, the Noise Visibility Function (NVF) of Voloshynovskiy [5] the complex image contrast of Peli [2] and more generally the idea of relative noise or Normalised Root Mean Square Error (NRMSE) [6]. Recently the close connection between perceptual masking of image watermarks and SSIM has been demonstrated [7]. Laparra's analysis for two image databases LIVE and TID2008 (figs 6 & 7 in [4]) shows that the Divisive Normalisation outperforms SSIM in terms of correlation with Difference Mean Opinion Scores (DMOS) of image distortion. Of the measures investigated only the VIF [8] slightly outperforms Laparra's divisive norm, and VIF achieves this with a significantly more computational, multi-channel HVS model.

Both the concept and the computation of SSIM can be replaced by the far simpler concept and computation of a universal noise visibility function (NVF) or **Dissimilarity Quotient (DQ):**

$$\boxed{D = \sqrt{\frac{1 - S_V}{2}} = \frac{\sigma_-}{\sqrt{\sigma_+^2 + \sigma_-^2}}} \tag{8a}$$

Here we have taken the liberty of naming the quantity **DQ** so as to distinguish the notational efficiency afforded by the symmetric formulation.

## Discussion

Applying Occam's Razor to our current understanding of SSIM inevitably leads to a concept like Dissimilarity Quotient with equivalent explanatory power and a simple, tractable, and perceptually pleasing mathematical formulation.

## Notes

### Avoidance of zero division
SSIM, NRMSE and Laparra's divisive norm suffer from potential blow-up caused by a zero denominator. The problem is conventionally addressed by adding a small regularising constant to the denominator in each case.

### Aggregation/Pooling
The spatially localised value of DQ can be aggregated into a single, overall value representing a complete image measurement. SSIM (quadratically related to DQ) typically uses spatial averaging which corresponds to Minkowski pooling of DQ with exponent of 2. Laparra's optimisation gave a spatial exponent of 2.2. However, given that a exponent of 1 corresponds to a more perceptually linear effect, it may be appropriate for DQ. Nevertheless, optimal spatial pooling remains to be determined.

### Tuning, Frequency Response, Scale invariant Measures
Models of image quality which more closely correspond to human observer DMOS ratings must inevitably be tuned to the specific HVS responses applied under specific viewing conditions and geometry. However many researchers are really looking for a robust image quality measure that applies for a large (and perhaps unspecified) range of viewing conditions and geometry. This means that such a measure must be gain, scale, and rotation invariant at the very least. The dissimilarity quotient fits the criteria, as do suitable invariant transforms of the image (applied before the variances are computed) such a gradient magnitude, Riesz transform and Laplacian. Of these only the Riesz transform maintains a neutral frequency effect and hence does not enhance high spatial frequency features.

### Extensions
Virtually all the proposed extensions to SSIM (e.g. gradient, Riesz transform, gradient magnitude, wavelet, multi-scale, curvelet, DCT, FFT, multi-channel HVS, etcetera) can be applied directly to the dissimilarity quotient.

### Previous work
Several other researchers have reached similar conclusions about SSIM versus *normalised* Mean Square Error over the last half dozen years. They have performed detailed comparisons and analyses, whereas we have tried to be more succinct than is probably possible.

## Minimal References